\pdfoutput=1

\documentclass[11pt]{article}

\usepackage[final]{acl}
\usepackage{times}
\usepackage{latexsym}
\usepackage[T1]{fontenc}
\usepackage[utf8]{inputenc}
\DeclareUnicodeCharacter{2713}{\checkmark}
\usepackage{microtype}
\usepackage{inconsolata}
\usepackage{graphicx}
\usepackage{amsmath, amssymb}
\usepackage{algorithm}
\usepackage{algpseudocode}
\usepackage{booktabs}
\usepackage{multirow}
\usepackage{url}
\usepackage{float}
\usepackage{tabularx}
\usepackage{enumitem}
\usepackage{longtable}
\usepackage{makecell}
\usepackage{setspace}
\usepackage{tcolorbox}
\newcommand{\yoni}[1]{\textcolor{red}{[Yoni]~#1}}

\title{Forget What You Know about LLMs Evaluations - LLMs are Like a Chameleon}

\author{
  Nurit Cohen-Inger\textsuperscript{1}, 
  Yehonatan Elisha\textsuperscript{2}, 
  Bracha Shapira\textsuperscript{1}, 
  Lior Rokach\textsuperscript{1}, 
  Seffi Cohen\textsuperscript{3} \\
  \\
  \textsuperscript{1}Ben Gurion University \quad
  \textsuperscript{2}Tel Aviv University \quad
  \textsuperscript{3}Harvard University
}

\date{}

\begin{document}
\maketitle

\begin{abstract}
Large language models (LLMs) often appear to excel on public benchmarks, but these high scores may mask an overreliance on dataset-specific surface cues rather than true language understanding. We introduce the \textbf{Chameleon Benchmark Overfit Detector (C-BOD)}, a meta-evaluation framework designed to reveal such overfitting. C-BOD systematically rephrases benchmark inputs via a parameterized transformation that preserves semantic content and labels, enabling the detection of performance degradation indicative of superficial pattern reliance.
We conduct extensive experiments across two datasets, three rephrasing models, and multiple distortion levels, evaluating 32 state-of-the-art LLMs. On the MMLU benchmark, C-BOD reveals an average performance drop of 2.75\% under modest rephrasings, with over 80\% of models exhibiting statistically significant differences. Notably, higher-performing models and larger LLMs tend to show greater sensitivity, suggesting a deeper dependence on benchmark-specific phrasing.
Due to its dataset and model-agnostic design, C-BOD can be easily integrated into evaluation pipelines and offers a promising foundation for overfitting mitigation strategies. Our findings challenge the community to look beyond leaderboard scores and prioritize resilience and generalization in LLM evaluation. Our code and benchmark datasets are available
at: \url{https://github.com/nuritci/cbod}
\end{abstract}

\section{Introduction}
\begin{figure}[ht!]
    \centering
    \includegraphics[width=0.95\linewidth]{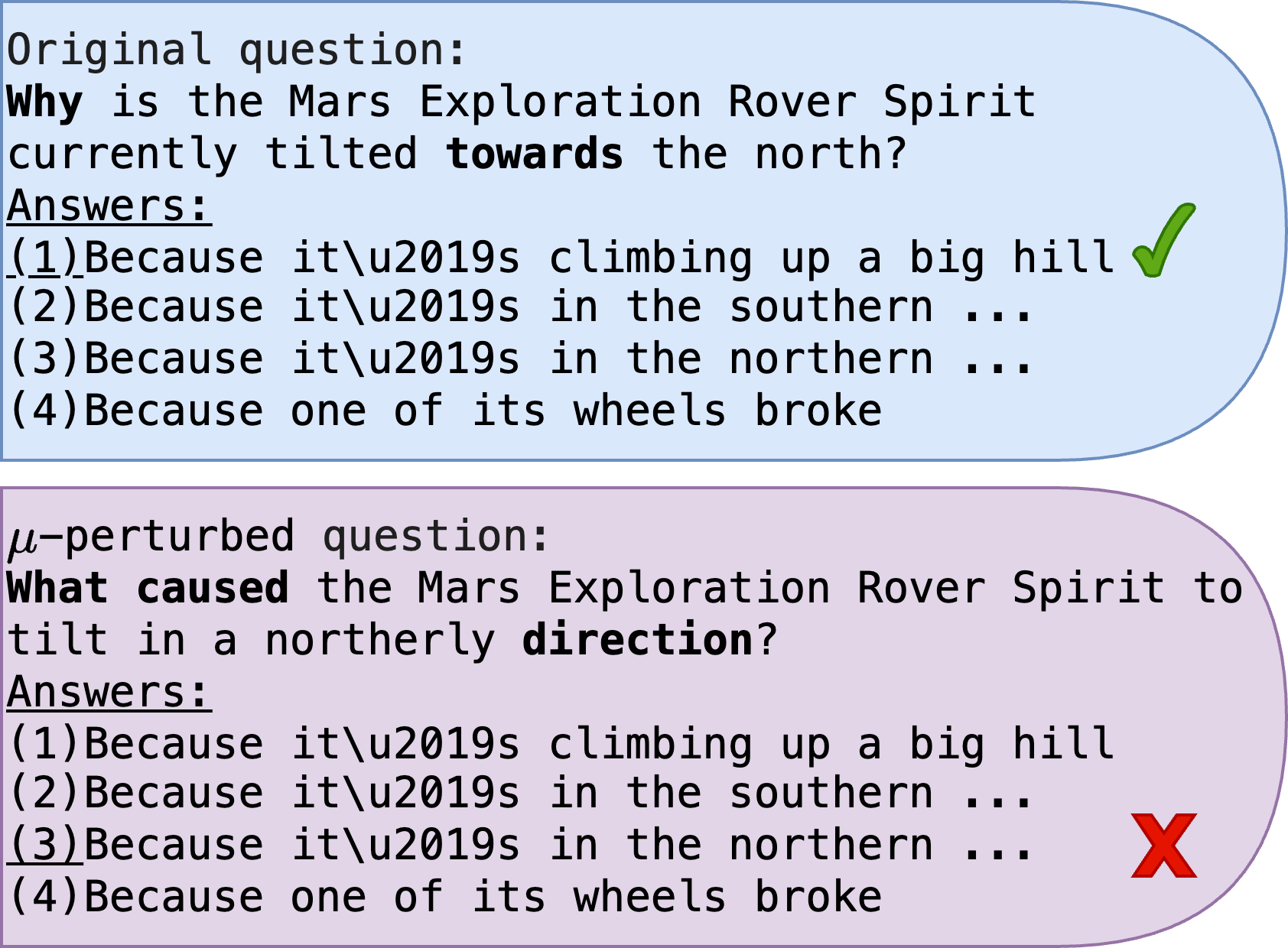} %
    \caption{An example demonstrating the C-BOD method. The original question (top) is perturbed (bottom) while preserving the semantic meaning and correct answer options.  The model correctly answers the original question but fails on the perturbed version, suggesting potential overfitting. Changes in the perturbed question are highlighted in bold.}
    \label{fig:mu_transform_examples}
\end{figure}

Large Language Models (LLMs) have achieved impressive results on a wide range of NLP tasks \cite{chang2024survey}. Consequently, hundreds of benchmarks have been established to track progress and evaluate model capabilities \cite{lu2024comprehensive, liang2022holistic}. However, the rapid proliferation of LLMs and the frequent use of public leaderboards raise concerns about the robustness of these evaluation practices \cite{castillo2024beyond}. Specifically, as benchmark data becomes more widely recognized, models may learn to exploit surface patterns or spurious correlations, rather than exhibit genuine language understanding. This issue can lead to deceptively high scores that do not reflect true progress.
In this paper, we examine whether LLMs rely excessively on benchmark-specific cues potentially overfitting to the patterns inherent in widely published evaluation benchmarks and explore systematic methods to detect and mitigate this behavior. In other words, are LLMs prone to overfitting on popular benchmarks, and what underlying factors contribute to this phenomenon?
To answer this question, we introduce the Chameleon Benchmark Overfit Detector (C-BOD). This framework reveals how heavily a model depends on the exact wording or structure of a test set. By introducing controlled textual distortions to benchmark prompts at varying intensities (defined by a distortion parameter $\mu$), as demonstrated in Figure \ref{fig:mu_transform_examples}, our method exposes whether strong performance derives from reliance on superficial patterns. Notably, our framework requires only the evaluation set, without accessing the model’s training data or architecture. Unlike conventional leaderboards that solely track performance, our meta-evaluation framework acts as a safeguard ensuring that high scores do not stem from superficial memorization of benchmark cues.

\paragraph{Our Contributions:}

\begin{enumerate}[leftmargin=1.2em, itemsep=0.5em, topsep=0em]
    \item \textbf{A Robust Framework for Detecting Benchmark Overfitting.} 
    We present a framework that computes the performance difference $\Delta_{\mu}$ between original and perturbed prompts and confirms its statistical significance, ensuring that observed differences indeed indicate overfitting rather than chance variations.

    \item \textbf{New Insights into LLM Behavior.} 
    Our analysis reveals that larger models and those with higher baseline performance are often more sensitive to perturbations, suggesting a deeper reliance on benchmark-specific phrasing.
    
    \item \textbf{Extensive Empirical Validation.} 
    We apply our method to a diverse collection of 32 leading LLMs from various families, architectures, and parameter sizes. Utilizing modest textual distortions generated by three different rephrasing models for enhanced robustness, our analysis reveals statistically significant performance degradation in over 80\% of the evaluated models, providing strong empirical evidence of widespread benchmark overfitting. Similar trends were observed in our evaluation on the GPQA dataset, in which we also show an ablation over different distortion levels ($\mu$). 

    \item \textbf{Open Resources for the Community.} 
    To facilitate further research and promote robust evaluation, we publicly release the rephrased versions of the widely used MMLU and GPQA evaluation sets, along with our reproducible code. These resources enable the community to readily adopt more robust, surface-invariant tests for reliable LLM assessment and provide a foundation for developing mechanisms to mitigate benchmark overfitting.

\end{enumerate}

\section{Related Work}
\label{sec:related_work}
\subsection{Benchmark Datasets and Evaluation Suites}

LLMs have achieved impressive results on many benchmarks. This success has driven the development of comprehensive evaluation suites such as BIG-Bench \cite{srivastava2022beyond} and HELM \cite{liang2022holistic}. The MMLU benchmark set \cite{hendrycks2020measuring} evaluates question answering across 57 subjects, including STEM, humanities, and social sciences, while \cite{bing_zhang__2024} introduced 25 enterprise-focused datasets covering domains like finance, legal, cybersecurity, and climate sustainability for tasks such as classification, NER, and summarization. Another recent resource, JUDGE-BENCH \cite{anna_bavaresco__2024}, comprises 20 NLP datasets that assess models against human judgments.  GPQA benchmark \cite{rein2024gpqa} is  tasked with evaluating reasoning models. We focus on MMLU and GPQA because of their widespread adoption\footnote{https://klu.ai/glossary/gpqa-eval} and comprehensive domain coverage  \cite{wang2024mmlu}. 

\subsection{Overfitting in LLMs}
While these benchmarks have been critical for comparing new models' versions, recent studies warn that publicly released evaluation sets can become less reliable over time due to overexposure and memorization \cite{yuan_yu__2024, chang2024survey}. In some cases, LLMs learn superficial patterns specific to well-known datasets, boosting performance without reflecting genuine semantic or conceptual understanding. \cite{kiela2021dynabench} further emphasizes the need for continuously refreshing benchmarks to ensure real progress in language understanding. For example, OpenAI's GPT models have shown steady improvement on MMLU: GPT-3 achieved approximately 43\% accuracy in 2020 \cite{brown2020language}, rising to nearly 70\% with GPT-3.5 in 2022, and reaching 86\% with GPT-4 in 2023 \cite{koubaa2023gpt}. 

Memorization in LLMs has been widely studied \cite{kiyomaru2024comprehensive, biderman2024emergent}, with larger models especially prone to retaining training data verbatim \cite{carlini2022quantifying}. This phenomenon can inflate performance metrics while obscuring genuine model capabilities. Moreover, several works highlight training-set contamination, where test samples appear exactly or as near-duplicates in the training data, as another crucial form of overfitting \cite{deng2023investigating, yao2024data}, leading to overly optimistic performance estimates \cite{yang2023rethinking}. Training Data Contamination refers to the presence of test data, or near duplicates, within the training set \cite{deng2023investigating, yao2024data}. Contamination renders evaluation unreliable, as the model has effectively already seen the "test" data, leading to overly optimistic performance estimates \cite{yang2023rethinking}.
Benchmark/Prompt Structure Overfitting is a more subtle form that arises when LLMs learn to exploit superficial cues or patterns specific to a benchmark's format or the structure of evaluation prompts, even without memorizing the exact content \cite{yuan_yu__2024}. This can lead to overestimated generalization ability, as the model's performance becomes dependent on the specific benchmark artifacts rather than true language understanding.  This type of overfitting is the focus of our work.

\subsection{Addressing the Gap}
Current methods largely overlook the crucial problem of overfitting to benchmark-specific artifacts, which can significantly misrepresent an LLM's true capabilities and hinder the development of robust and generalizable models.  Our work addresses this gap by introducing a novel method to quantify an LLM's reliance on benchmark prompt structure.  We systematically apply controlled distortions to evaluation prompts, for example, by replacing synonyms or altering word order and measure the resulting performance degradation.  This approach, which does not require access to training data, provides a direct measure of vulnerability to prompt structure and a robust means of diagnosing and mitigating this critical form of overfitting.

\section{Method}
\label{sec:method}
\begin{figure}[ht]
    \centering
    \includegraphics[width=0.52\textwidth]{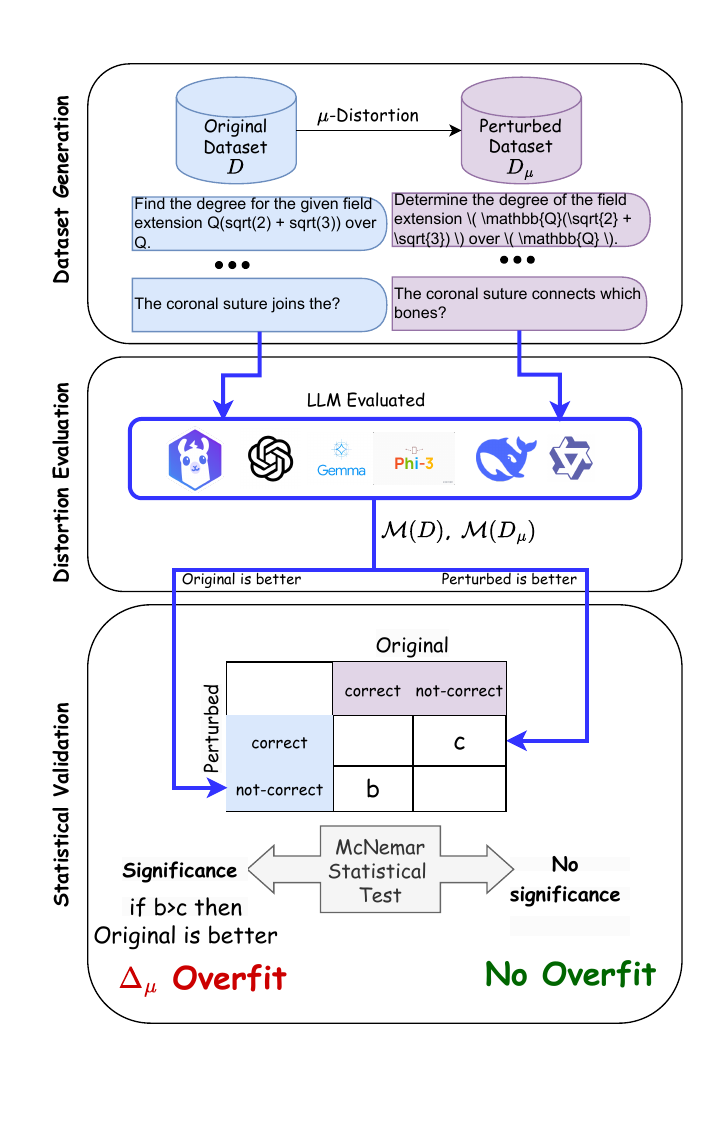}
    
    \caption{High-level pipeline of our parametric approach. The original dataset \(\mathcal{D}\) is passed through the distortion operator \(T_{\mu}\) to form \(\mathcal{D}_{\mu}\). Both sets are evaluated by an LLM, and differences in performance are used to statistically quantify overfitting.}
    \label{fig:method_flow}
\end{figure}

Let \(\mathcal{D}\) denote a benchmark dataset with N samples, and \(\mathcal{E}\) an LLM to be evaluated with respect to a given performance function \(\mathcal{M}\).
Our goal is to detect whether \(\mathcal{E}\) exhibits overfitting to \(\mathcal{D}\).
Figure~\ref{fig:method_flow} provides an overview of our proposed method, Chameleon Benchmark Overfit Detector (C-BOD).
C-BOD employs a rephrasing transformation to generate a perturbed dataset from \(\mathcal{D}\), evaluates on both the original and perturbed datasets, and applies a statistical test to assess whether performance discrepancies indicate overfitting. The following subsections detail each component of C-BOD.

\subsection{C-BOD rephrased dataset generation}
To systematically introduce textual variations, C-BOD utilizes a rephrasing tool, denoted as \({T}\), which uses as a distortion operator to generate a perturbed dataset \(\mathcal{D}_{\mu}\) from \(\mathcal{D}\). This operator is parameterized by \(\mu\), which controls the extent of textual modification, ranging from low (e.g., 0.1 for minimal changes like synonym substitution) to moderate (e.g., 0.5 for rewording and sentence fragment reordering) and high (e.g., 1.5 for aggressive modifications such as question reformulation). Specifically, for an LLM, \(\mu\) corresponds directly to the temperature used during generation, influencing the degree of variation in rephrased outputs. We define:

\[
T_{\mu} : \mathcal{X} \rightarrow \mathcal{X'}
\]

Given a prompt \(x_i\), the distortion operator produces a perturbed prompt \(x_i' = T_{\mu}(x_i)\). The perturbed dataset is then constructed as:

\[
\mathcal{D}_{\mu} \;=\; \left\{\,\bigl(x_i',\,y_i\bigr)\;\middle|\;(x_i,y_i)\in \mathcal{D}\right\}
\]

Although each pair \((x_i', y_i)\) in the perutbed dataset remains semantically equivalent to \((x_i, y_i)\) in the original dataset, the textual variations introduced by \(T_{\mu}\) can disrupt purely memorized mappings from surface patterns to correct labels.
This step presented in Lines 5-6 of Algorithm~\ref{alg:cbod}.

\subsection{Evaluating the Impact of Distortion} 
To assess the impact of distortion, we evaluate \(\mathcal{E}\) using a performance function, \(\mathcal{M}\). This function evaluates \(\mathcal{E}\) based on a given ground truth \( y_i \), considering two versions of an input: the original \( x_i \in \mathcal{D} \) and the perturbed version \( x_i' \in \mathcal{D}_{\mu} \), where \( i \) denotes the index of a sample in the dataset. Specifically, \(\mathcal{M}\) is a boolean function that takes as input the model \(\mathcal{E}\) and two data pairs, \((x_i, y_i)\) and \((x_i', y_i)\), and returns whether the model performs better on the original input than on the perturbed one. The function is formulated as follows:  
\[
\mathcal{M}(\mathcal{E}, (x_i, y_i), (x'_i, y_i) )=
\begin{cases}
    1, & \text{if } P(\mathcal{E}, x_i, y_i) \\
       & \quad > P(\mathcal{E}, x'_i, y_i), \\
    0, & \text{otherwise.}
\end{cases}
\]

where \( P(\mathcal{E}, x, y) \) represents the performance score of model \(\mathcal{E}\) on input \( x \) with reference to ground truth \( y \). This formulation is designed to be generalizable across different evaluation metrics and natural language understanding (NLU) tasks.  
The performance difference between the original set and the perturbed set is then calculated as:
\begin{equation}
\Delta_{\mu} = b =\sum_{i=0}^{N} \mathcal{M}(\mathcal{E}, (x_i, y_i), (x_i', y_i))
\end{equation}
The performance difference between the perturbed set and the original set is then calculated as:
\begin{equation}
c =\sum_{i=0}^{N} \mathcal{M}(\mathcal{E}, (x_i', y_i), (x_i, y_i))
\end{equation}

A large positive \(\Delta_{\mu}\) indicates a significant performance decline due to textual perturbations, suggesting that \(\mathcal{E}\) may be overly reliant on surface-level patterns rather than exhibiting robust generalization. Notably, this approach remains metric-agnostic, making it applicable to a wide range of evaluation measures. This step presented in Lines 7-8 of Algorithm~\ref{alg:cbod}.
\subsection{Statistical Validation}
To assess the statistical significance of performance differences, we employ McNemar’s test \cite{mcnemar1947note}, which is specifically designed for paired data. This test evaluates whether the discrepancies between two related sets of classification outcomes, correct and incorrect predictions, are significant. In our context, McNemar’s test is well-suited for comparing each pair of samples \((x_i, y_i) \in D\) and \((x_i', y_i) \in D_\mu\), we record whether \(\mathcal{E}\) classifies them correctly and aggregate into b (original is better) and c (perturbed is better) as presented in Equation 1, Equation 2.
The McNemar statistic is then calculated as: 
\begin{equation}    
\chi^2 = \frac{(b - c)^2}{b + c} 
\end{equation}

We derive a \(p\)-value from the chi-squared distribution (with df=1, i.e., one degree of freedom), rejecting the null hypothesis if \(p < \alpha\). A significant result with \(b>c\) indicates a genuine performance difference due to the transformation, suggesting evidence of overfitting.
This step presented in Lines 10-19 of Algorithm~\ref{alg:cbod}.

\begin{algorithm}[ht!]
\caption{Chameleon Benchmark Overfit Detector}
\label{alg:cbod}
\begin{algorithmic}[1]
\Require 
\Statex $\mathcal{D}$: Original benchmark dataset of size $N$, 
\Statex $\mathcal{E}$: LLM,
\Statex $\mu$: Distortion parameter,
\Statex $T_{\mu}$: Transformation operator,
\Statex $\mathcal{M}$: Performance function (returns 1 if the first input is better, 0 otherwise),
\Statex $\alpha$: Significance level.

\State \textbf{C-BOD Computation:}

\State $b,c \leftarrow 0$
\State \(D_\mu \leftarrow \{\}\)
\For{each \(x_i \in \mathcal{D}\)}
    \State \(x'_i \leftarrow T_{\mu}(x_i)\)
    \State \(D_\mu \leftarrow D_\mu \cup\,   x_i'\)
    \State \(b \leftarrow b + \mathcal{M}(\mathcal{E}, (x_i, y_i), (x_i', y_i))\)
    \State \(c \leftarrow c + \mathcal{M}(\mathcal{E}, (x_i', y_i), (x_i, y_i))\)
\EndFor

\State \(\chi^2 \leftarrow \dfrac{(b-c)^2}{b+c}\)
\State \(p \leftarrow \text{p-value}(\chi^2, \text{df}=1)\)


\If{\(p < \alpha\)}
    \If{\(b > c\)}
        \State \(Overfit\_Flag \leftarrow True \)
    \Else
        \State \(Overfit\_Flag \leftarrow False \)
    \EndIf
\Else
    \State \(Overfit\_Flag \leftarrow False \)
\EndIf

\State \Return \(Overfit\_Flag\)
\end{algorithmic}
\end{algorithm}

\section{Experimental Setting}
\label{sec:experiments}

In this section, we describe the experimental setup used to evaluate our overfitting detection framework. We detail the benchmark dataset, the procedure for generating perturbed inputs, the LLMs under evaluation, implementation specifics, and the evaluation metrics.

\subsection{Dataset and Rephrasing Process}

Our experiments used two leading benchmark datasets: (1) \textbf{MMLU}~\cite{hendrycks2020measuring}: This benchmark spans 57 subjects, comprising 14,079 test samples and 1,540 validation samples. Its broad coverage makes it a standard choice for evaluating general knowledge and assessing overfitting to canonical prompt formats. It is distributed under the MIT License, allowing free use and modification.
(2) \textbf{GPQA}~\cite{rein2024gpqa}: Introduced in 2024, this smaller yet challenging benchmark includes 546 multi-step reasoning samples designed to test logical inference. Its complex, less common question structures help reduce the risk of training data contamination, making it a valuable complement to MMLU for overfitting analysis. Detailed results and analysis on GPQA are provided in Appendix \ref{sec:appendix GPQA}.

We generate a perturbed version of the original dataset to probe overfitting, following the methodology described in Section~\ref{sec:method}. We used \texttt{DeepSeek~3} to create the transformed version of each question and generate the perturbed dataset \(\mathcal{D}_{1.0}\) using \(\mu = 1.0\) (the default temperature parameter for DeepSeek), and \texttt{Claude~3.5~Haiku} to generate the perturbed dataset \(\mathcal{D}_{0.5}\) using \(\mu = 0.5\) (the default temperature parameter for Claude). For the GPQA we used \texttt{GPT-4o-mini} to generate the perturbed datasets \(\mathcal{D}_{0.5}\), \(\mathcal{D}_{1.0}\), \(\mathcal{D}_{1.5}\), using \(\mu \in \{0.5, 1.0, 1.5\}\). 

These perturbations include synonym substitutions, sentence reordering, and the insertion of distractor phrases, while preserving the original semantic meaning and correct answers. The perturbed datasets, denoted by \(\mathcal{D}_{\mu}\), is released alongside our code for reproducibility.

\subsubsection{Evaluation of Rephrasing Quality}

To ensure the quality of rephrasing in the C-BOD framework, we implemented a multi-step evaluation approach to maintain the semantic integrity of the original prompts and avoid confounding overfitting assessments. This process included three main validation steps: (1) Cosine Similarity Analysis, (2) Semantic Equivalence Verification, and (3) Iterative Human Audits. In what follows, we briefly describe these steps.

\paragraph{Cosine Similarity Analysis} We measured semantic alignment using Sentence-BERT~\cite{reimers2019sentence}, confirming high alignment across perturbations: 
\begin{itemize} [leftmargin=1em]

\item \emph{MMLU 0.5 (Claude)}: \\Mean = 0.829, Median = 0.864.
\item \emph{MMLU 1.0 (DeepSeek)}:\\ Mean = 0.883, Median = 0.922. 
\item \emph{GPQA 0.5 (GPT)}:\\ Mean = 0.954, Median = 0.969. 
\item \emph{GPQA 1.0}:\\ Mean = 0.947, Median = 0.967. 
\item \emph{GPQA 1.5}:\\ Mean = 0.942, Median = 0.961.
\end{itemize}
\paragraph{Semantic Equivalence Verification} In addition to cosine similarity, we employed a reasoning model \texttt{GPT-o3} to independently verify the retention of original intent. Approximately 1-2\% of the rephrasings exhibited semantic errors or insufficient similarity, requiring further manual correction and refinement before inclusion in the final evaluation set. This additional verification reduced the risk of subtle semantic drift, ensuring a high-quality perturbation set.

\paragraph{Iterative Human Audits} To ensure the accuracy and semantic consistency of the perturbed datasets, we conducted a comprehensive manual audit as the final validation step. This process specifically targeted prompts with a cosine similarity score below 0.7 or those that received a negative judgment from the automated evaluation. These prompts were iteratively refined through API adjustments until they met the desired fidelity thresholds. Overall, this approach manually covered approximately 20\% of the MMLU dataset and 50\% of the GPQA dataset, supplementing the earlier automated checks that covered 100\% of both datasets. This multi-step approach significantly enhances the reliability of our overfitting assessments by ensuring precise, controlled textual variations.

\subsection{Models Under Evaluation}
\label{subsec:models_eval}

Table~\ref{tab:models_overview} in Appendix \ref{sec:appendix models} provides an overview of the LLMs evaluated in our experiments. Our study covers a diverse set of architectures and parameter scales ranging from 1B to 27B parameters. This broad selection enables an in-depth analysis of how both architectural choices and model scale affect robustness to prompt perturbations.

\subsection{Implementation Details}

All experiments were executed under standardized conditions to ensure reproducibility and fair comparisons:
\begin{enumerate}[nosep, label=(\arabic*)]
    \item \textbf{Inference Environment:} All open-weight models were accessed via the HuggingFace transformers library using RTX 6000 GPU. 
    \item \textbf{Dataset Rephrasing Prompt:}  
    We instruct the rephrasing tool using the prompt detailed in Appendix \ref{sec:appendix prompt}. 
    
    \item \textbf{Query Prompt:}  
    For every query, we construct a standardized input by prompting a fixed instruction to the original benchmark dataset question. Importantly, the multiple-choice options remain identical between the original and the rephrased forms. The fixed instruction is: 
    \begin{tcolorbox}
    “Select the best answer from the given options. Respond with only the letter corresponding to the correct choice.  Question: \{question\}”
    \end{tcolorbox}

\end{enumerate}

\subsection{Evaluation Metrics}

We assess model performance by comparing the original dataset, \(\mathcal{D}\), with its perturbed counterpart, \(\mathcal{D}_{0.5}\), \(\mathcal{D}_{1.0}\), using the following metrics:

\textbf{Correct Predictions and Accuracy:} For each dataset, we report the number of correct answers and the corresponding accuracy, defined as
    \[
    \text{Accuracy} = \frac{\#\text{Correct Predictions}}{\#\text{Total Samples}}.
    \]
    
\textbf{Absolute and Percentage Performance Difference:} The absolute difference in the number of correct answers between \(\mathcal{D}\) and \(\mathcal{D}_{0.5}\) is denoted by \(\Delta_{0.5}\); we also report the relative difference.
 \textbf{Statistical Significance:} McNemar’s test is applied on the paired predictions to determine whether the performance gap is statistically significant (\(p < 0.05\)).



\section{Results}
\label{sec:results}

\subsection{Open Weight Models - Overall Performance}
As shown in Table \ref{tab:results}, the majority of models (22 out of 27 for \(D_{0.5}\) and 20 out of 27 for \(D_{1.0}\)) exhibit a noticeable drop in performance on the rephrased test set compared to the original, supporting our hypothesis that many LLMs are sensitive to prompt structure. Notably, smaller models like \texttt{Llama 1B} and \texttt{Llama 3B} maintained relatively stable accuracy, suggesting they are less prone to overfitting, potentially due to their more limited capacity for memorizing superficial patterns.

We also observed that models with lower baseline accuracy tend to show statistically insignificant differences, likely because their initial performance leaves less room for detectable degradation. Importantly, McNemar’s test confirmed that the observed performance drops in most models were statistically significant (\(p < 0.05\)), reinforcing the reliability of our method.

Across all evaluated models, the average drop in accuracy for \(D_{0.5}\) was 2.15\%, which increased to 2.75\% when considering only models with statistically significant differences. For \(D_{1.0}\), the average drops were 1.87\% overall and 2.78\% for models with significant performance changes, underscoring the broader impact of stronger perturbations.

\begin{table*}[h!]
\centering
\caption{Comparison of LLM performance on the original and perturbed MMLU datasets. Models are sorted by parameter count (ascending).}
\label{tab:results}
\footnotesize
\begin{tabular}{
    l      
    c      
    c      
    ||
    c      
    c      
    c      
    l      
    ||
    c      
    c      
    c      
    l      
}
\toprule
\makecell{\textbf{Model} \\ \textbf{Name}} &
\makecell{\textbf{Par} \\ \textbf{(B)}} &
\makecell{\textbf{$\mathcal{D}$} \\ \textbf{Accuracy}} & 
\multicolumn{4}{c}{\textbf{Claude (0.5)}} &
\multicolumn{4}{c}{\textbf{DeepSeek (1.0)}} \\
\cmidrule(lr){4-7} \cmidrule(lr){8-11}
& & 
&
\makecell{\textbf{$\mathcal{D}_{0.5}$} \\ \textbf{Accuracy}} &
\makecell{\textbf{\#} \\ \textbf{\(\Delta_{0.5}\)}} &
\makecell{\textbf{\%} \\ \textbf{\(\Delta_{0.5}\)}} &
\makecell{\textbf{Better} \\ \textbf{0.5}} &
\makecell{\textbf{$\mathcal{D}_{1.0}$} \\ \textbf{Accuracy}} &
\makecell{\textbf{\#} \\ \textbf{\(\Delta_{1.0}\)}} &
\makecell{\textbf{\%} \\ \textbf{\(\Delta_{1.0}\)}} &
\makecell{\textbf{Better} \\ \textbf{1.0}} \\
\midrule

Gemma-2B & 2 & 47.28 & 46.47 & 124 & 1.86 & Original & 46.54 & 104 & 1.56 & Original \\
Gemma-4B & 4 & 66.73 & 54.78 & 222 & 2.80 & Original & 65.07 & 234 & 2.49 & Original \\
Gemma-7B & 7 & 58.05 & 55.18 & 127 & 1.61 & Original & 57.18 & 123 & 1.50 & Original \\
Gemma-12B & 12 & 68.71 & 68.77 & 320 & 3.20 & Original & 66.38 & 328 & 3.39 & Original \\
Gemma-27B & 27 & 73.16 & 70.80 & 356 & 3.45 & Original & 70.34 & 397 & 3.85 & Original \\
Llama-1B & 1 & 28.11 & 28.09 & 3 & 0.08 & Not Sig & 27.00 & -3 & -0.08 & Not Sig \\
Llama-3B & 3 & 56.12 & 55.49 & 89 & 1.13 & Not Sig & 56.74 & -49 & -0.62 & Not Sig \\
Llama-8B & 8 & 45.93 & 45.21 & 102 & 1.58 & Original & 44.68 & 121 & 1.92 & Original \\
Mistral-7B & 7 & 57.48 & 56.33 & 163 & 2.01 & Original & 56.31 & 128 & 1.59 & Original  \\
Mistral-8B & 8 & 68.05 & 65.94 & 298 & 3.10  &  Original & 66.01 & 262 & 2.95 & Original \\
Phi-3.8B & 3.8 & 56.42 & 56.01 & 57 & 0.72 & Not Sig & 56.31 & -44 & -0.56 & Not Sig \\
Phi-14.7B & 14.7 & 76.77 & 74.69 & 294 & 2.72 & Original & 74.79 & 246 & 2.28 & Original \\
Phi-14.7B-Reasoning & 14.7 & 73.94 & 72.05 & 303 & 2.90 & Original & 72.93 & 142 & 1.36 & Original \\
Qwen-0.6B & 0.6 & 39.29 & 39.20 & 12 & 0.22 & Not Sig & 39.48 & -33 & -0.60 & Not Sig \\
Qwen-1.5B & 1.5 & 38.23 & 36.46 & 249 & 4.63 & Original & 35.41 & 151 & 2.94 & Original \\
Qwen-1.7B & 1.7 & 52.98 & 51.94 & 147 & 1.97 & Original & 52.20 & 91 & 1.22 & Not Sig \\
Qwen-3B & 3 & 41.49 & 41.46 & 5 & 0.09 & Not Sig & 40.76 & 97 & 1.66 & Not Sig \\
Qwen-4B & 4 & 66.97 & 64.27 & 381 & 4.04 & Original & 65.07 & 234 & 2.49 & Original \\
Qwen2.5-7B-Instruct & 7.0 & 58.72 & 56.92 & 253 & 3.06 & Original & 48.37 & 180 & 2.58 & Original \\
Qwen3-8B & 8.0 & 69.73 & 67.83 & 267 & 2.72 & Original & 67.82 & 251 & 2.56 & Original \\
Qwen3-14B & 14.0 & 66.86 & 65.87 & 139 & 1.48 & Original & 66.11 & 17 & 0.18 & Not Sig \\
Yi-6B & 6 & 63.19 & 60.63 & 361 & 4.06 & Original & 60.55 & 374 & 4.20 & Original \\
Yi-9B & 9 & 66.38 & 65.17 & 170 & 1.82 & Original & 65.20 & 165 & 1.77 & Original \\
Apollo-7B & 7 & 67.81 & 65.42 & 337 & 3.53 & Original & 64.96 & 401 & 4.20 & Original \\
GLM-9B & 9 & 68.71 & 66.06  & 296  &  3.06 &  Original & 66.38 & 328 & 3.39 & Original \\
Starling-7B & 7 & 59.41 & 58.15 & 177 & 2.12 & Original & 58.09 & 185 & 2.21 & Original \\
Zephyr-7B & 7 & 56.25 & 54.92 & 194 & 2.45 & Original & 55.05 & 169 & 2.13 & Original \\
\bottomrule
\end{tabular}
\end{table*}

\subsection{Relationship Between Model Size and Overfit Detection}
\begin{figure}[ht]
    \centering
    \includegraphics[width=0.44\textwidth]{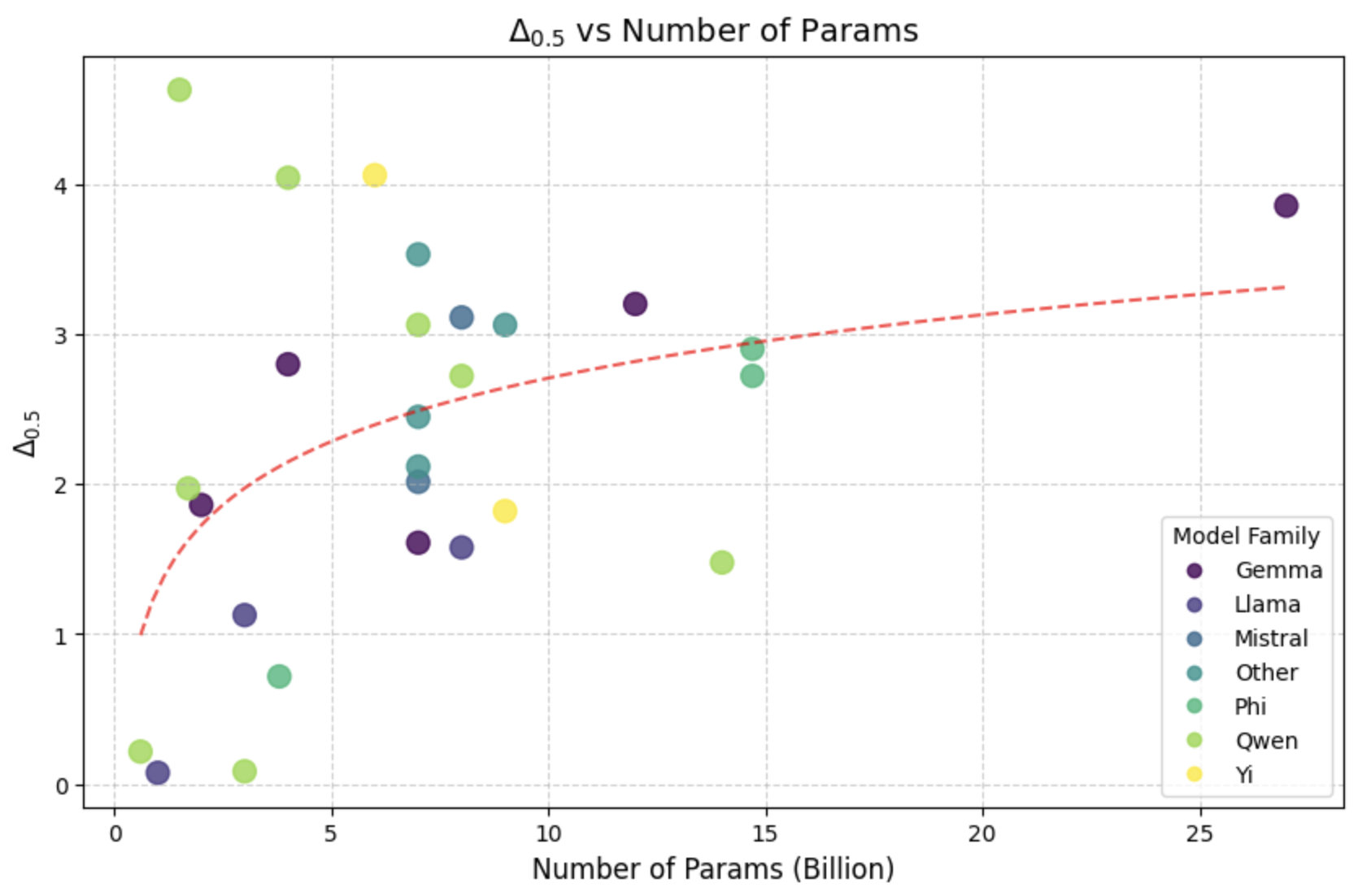}
    \caption{Scatter plot of the performance difference ($\Delta_{0.5}$) versus the number of model parameters. A logarithmic trendline is shown:\\ \(\Delta_{0.5} = 0.6090 \cdot \ln\bigl(\text{\# Params}\bigr) + 1.303\).}
    \label{fig:scatter_log_fit2}
\end{figure}

\begin{figure}[ht]
    \centering
    \includegraphics[width=0.44\textwidth]{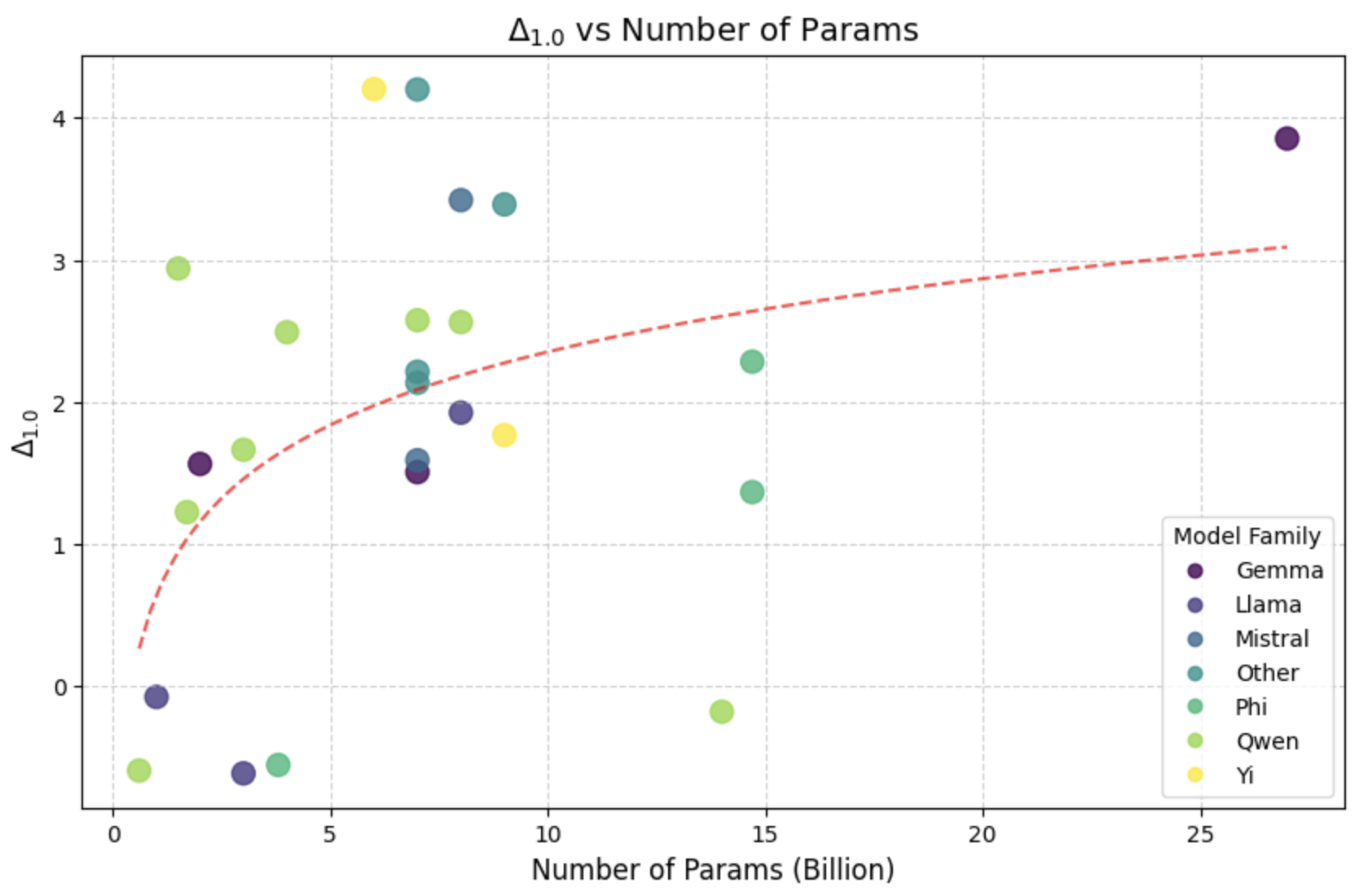}
    \caption{Scatter plot of the performance difference ($\Delta_{1.0}$) versus the number of model parameters. A logarithmic trendline is shown: \\ \(\Delta_{1.0} = 0.7433 \cdot \ln\bigl(\text{\# Params}\bigr) + 0.6406\).}
    \label{fig:scatter_log_fit}
\end{figure}

Figures \ref{fig:scatter_log_fit2}, \ref{fig:scatter_log_fit} illustrate the scatter plot of the percentage performance difference versus the number of parameters, with a red dashed line representing the logarithmic fit. The significant log-linear relationship indicates that the performance difference increases with model size in a logarithmic fashion, suggesting diminishing returns as the number of parameters grows.
The data reveals a positive trend: larger models tend to exhibit greater performance degradation under textual perturbations. For example, models in the Gemma family show a progressive increase in  \(\Delta_{1.0}\) with higher parameter counts. The dotted trend line further highlights this relationship.

\subsection{Relationship Between Model Accuracy and Overfit Detection}

\begin{figure}[h]
    \centering
    \includegraphics[width=0.44\textwidth]{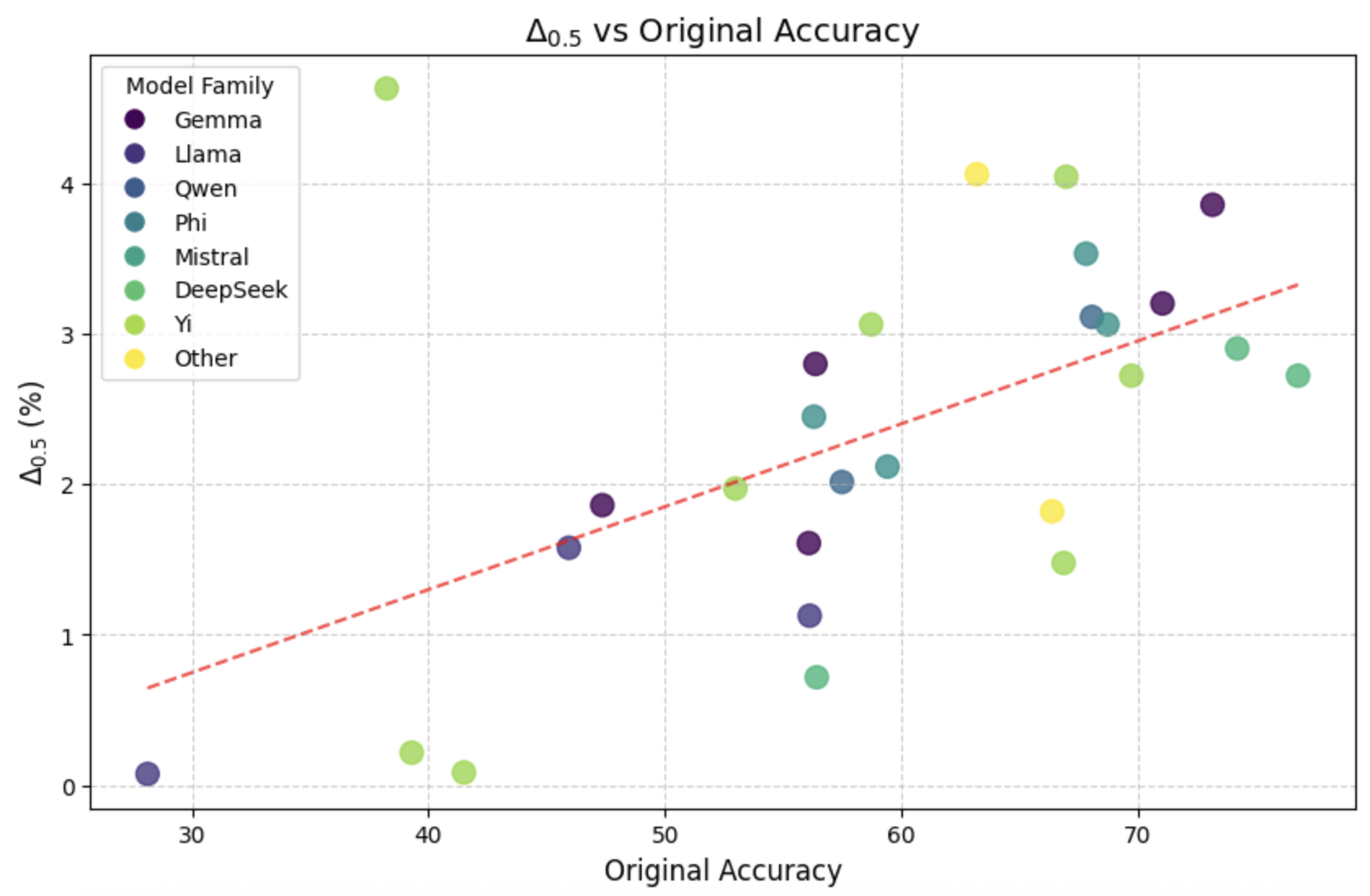}
    \caption{Scatter plot showing  \(\Delta_{0.5}\) for $\mu=0.5$ against the original accuracy of the model. Models within the same family are marked with the same color.}
    \label{fig:scatter_diff_acc}
\end{figure}

\begin{figure}[h]
    \centering
    \includegraphics[width=0.44\textwidth]{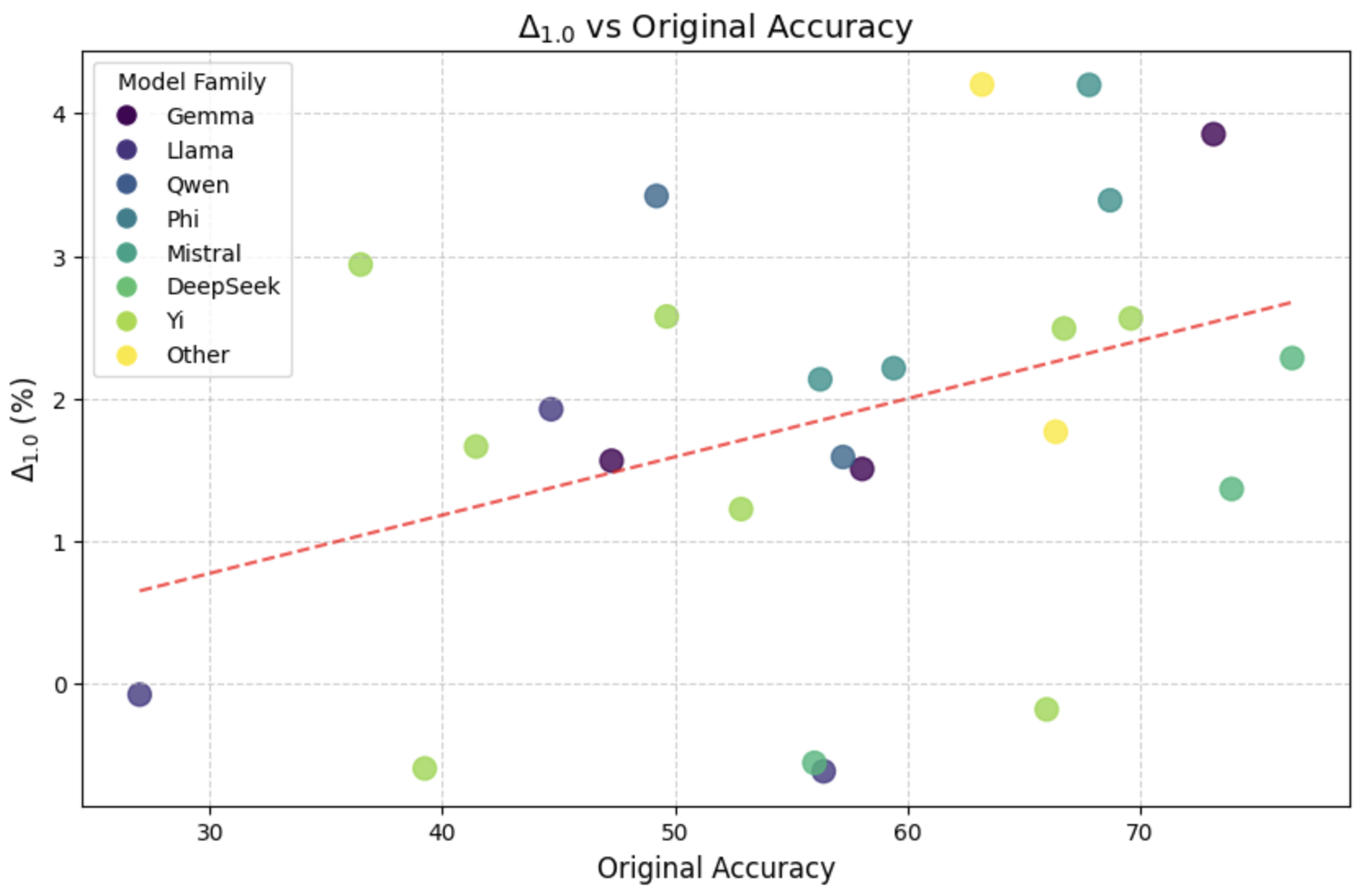}
    \caption{Scatter plot showing  \(\Delta_{1.0}\) for $\mu=1.0$ against the original accuracy of the model. Models within the same family are marked with the same color.}
    \label{fig:scatter_diff_acc1}
\end{figure}

Figures~\ref{fig:scatter_diff_acc}, \ref{fig:scatter_diff_acc1} examine the relationship between baseline accuracy on the original prompts and the corresponding percentage difference in performance when evaluated on rephrased inputs. The plot clearly indicates that models with higher original accuracy tend to experience larger declines when exposed to prompt perturbations. For example, models achieving over 60\% accuracy on the original set present the largest \(\Delta_{0.5}\), \(\Delta_{1.0}\) values, while models with lower baseline accuracy exhibit only minor, often statistically insignificant, differences.

This observation highlights a paradox in current LLM evaluation: models that perform exceptionally well on standard benchmarks may be capitalizing on dataset-specific cues rather than demonstrating robust language understanding. The positive correlation between original accuracy and  \(\Delta_{\mu}\) underscores the need to carefully interpret high benchmark scores, as they might mask underlying vulnerabilities to prompt variations.

\begin{table*}[h]
\centering
\caption{Examples of how rephrasing affects LLM performance, illustrating potential overfitting to specific phrasing in the original MMLU dataset. The table shows original and rephrased questions, along with an explanation of why the model's prediction changed. The examples are from Qwen3 (14B parameters).}
\scriptsize 
\renewcommand{\arraystretch}{1.1} 
\begin{tabular}{p{1.4cm}|p{3.3cm}|p{3.3cm}|p{6.4cm}}
\hline
\textbf{Subject} & \textbf{Original Question} & \textbf{Rephrased Question} & \textbf{Why the Model Was Wrong?} \\
\hline

Professional Law &
“If the defendant is \textbf{prosecuted} for the man’s murder, he will most \textbf{likely be found}...” &
“If the defendant is \textbf{charged} with the man’s murder, what is the \textbf{most probable} outcome?” &
In legal contexts, terms like “prosecuted” and “found guilty/not guilty” are tied to specific legal standards. The rephrased question is more open-ended, leading the model to discuss outcomes like plea bargaining instead of focusing on the legal verdict. \\
\hline

Moral Disputes &
“Of the following social \textbf{problems} that could result from a genetic supermarket, which does Singer think is the least \textbf{serious}?” &
“Which of the following social \textbf{issues} arising from a genetic supermarket does Singer consider to be the least \textbf{concerning}?” &
The word “problems” was changed to “issues,” altering the model’s interpretation. “Issues” can broaden the context of "problems", causing the model to incorrectly interpret which concerns are least serious. \\
\hline

College Chemistry &
“Which of the following statements is not a \textbf{reason} why tetramethylsilane is used as a 1H chemical shift reference?” &
“Which of the following statements does not \textbf{explain} why tetramethylsilane is used as a reference for 1H chemical shifts?” &
The model may have overfit to the structure of the original question, particularly the phrase “is not a reason why,” as it directly signals the correct retrieval path. The rephrased version, with slight syntactic adjustments disrupts this memorization, leading to incorrect retrieval. \\
\hline

World Religions &
“\textbf{When} did the first Jaina temples \textbf{appear}?.” &
“\textbf{At what point in time} were the initial Jaina temples \textbf{established}?” &
The rephrased question shifts key terms (“When” to “At what point in time”), obscuring historical framing. The LLM fails to map this modified phrasing to the original temporal context. \\
\hline
High-School Biology &
“Which of the following is \textbf{NOT} a characteristic of bacteria?” &
“Which of the listed options \textbf{fails to represent} a defining trait of bacterial organisms?” &
The explicit cue word “NOT” is replaced by the longer clause “fails to represent.”  
LLM appears to rely on a pattern like “Which … is not …” to flip polarity; when the negator is hidden inside a relative clause, the learnt template does not fire, so the model picks a true trait instead of the exception.\\
\hline
High-School Mathematics &
 “Positive integers $x,y$ satisfy $xy=56$, $x<y$, and $7$ divided by the \textbf{reciprocal of the larger integer} equals $4$. What is $x$?” &
 “$x,y$ are positive integers with product 56 and $x<y$. If \textbf{seven divided by the larger integer} results in 4, determine $x$.” &
The word “reciprocal” pins a template: $7 \times \frac{1}{y}=4 \Rightarrow y=\frac{7}{4}$.  
Replacing it with a looser “seven divided by the larger integer results in 4” flips the parse—The LLM treats “results in” like a remainder cue ($7\ \text{div}\ y = 4$), producing $y=1$ and thus the wrong $x$.\\
\hline
\end{tabular}
\label{tab:examples}
\end{table*}
These findings underscore the importance of evaluating LLMs under varied prompt formulations to ensure that improvements in benchmark performance reflect genuine advances in language understanding rather than overfitting.

\subsection{Closed-Source Models: Accuracy vs. Robustness Trends}
\label{subsec:closed_models}

We analyze the performance of proprietary GPT models (GPT-4o\footnote{https://openai.com/index/gpt-4o-system-card/?utm\_source=chatgpt.com}, GPT-4.1\footnote{https://openai.com/index/gpt-4-1/?utm\_source=chatgpt.com}, GPT-5\footnote{https://openai.com/index/gpt-5-system-card/?utm\_source=chatgpt.com})  under the C-BOD rephrasing evaluation. While these models achieve high accuracy on unperturbed MMLU prompts, they still show significant degradation under semantically equivalent prompt modifications (\(\mu = 0.5\)), as shown in Table \ref{tab:closed_models}.

\begin{table}[h]
\centering
\scriptsize
\begin{tabular}{lccccc}
\toprule
\textbf{Model} & \multicolumn{2}{c}{\textbf{Accuracy}} & \textbf{\#$\Delta_{0.5}$} & \%$\Delta_{0.5}$ & \textbf{Better} \\
\cmidrule(lr){2-3}
 & $\mathcal{D}$ & $\mathcal{D}_{0.5}$ &  &  &  \\
\midrule
GPT-4o          & 83.88\% & 81.46\% & 339 & 2.88 & Original* \\
GPT-4o-mini     & 74.96\% & 72.37\% & 364 & 3.46 & Original* \\
GPT-4.1-nano    & 69.68\% & 67.53\% & 302 & 3.09 & Original* \\
GPT-5-nano      & 86.63\% & 84.30\% & 326 & 2.68 & Original* \\
GPT-5-mini      & 90.56\% & 88.35\% & 310 & 2.44 & Original* \\
\bottomrule
\end{tabular}
\caption{Performance of closed-source GPT models on MMLU under prompt perturbation (\(\mu = 0.5\)). All results are statistically significant (marked with *) according to McNemar’s test (p < 0.05).}
\label{tab:closed_models}
\end{table}

\begin{figure}[h]
    \centering
    \includegraphics[width=0.85\linewidth]{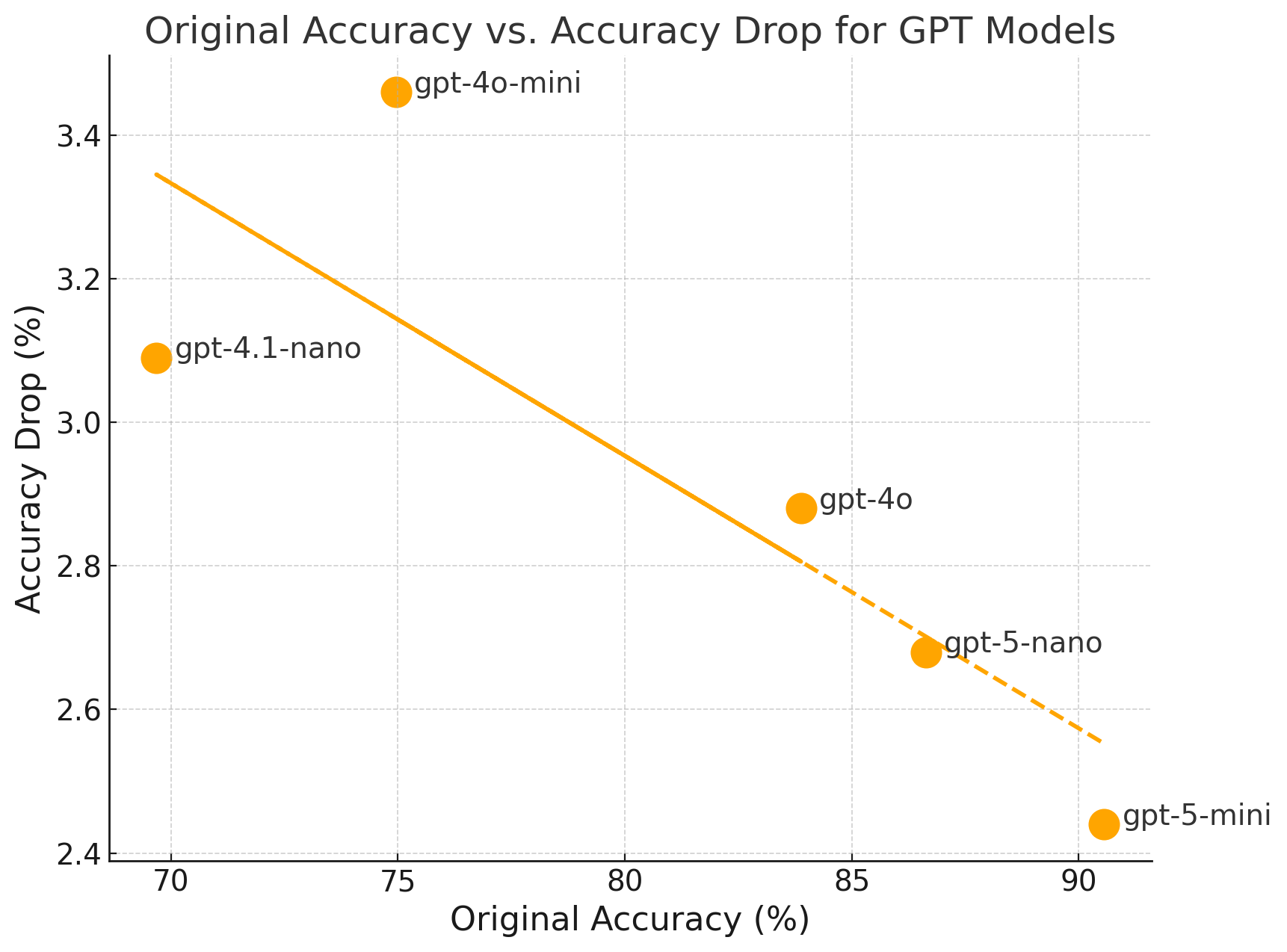}
    \caption{Accuracy vs. Drop for GPT models. Each point shows original accuracy and percentage drop after rephrasing.}
    \label{fig:gpt_accuracy_drop}
\end{figure}

\paragraph{Closed-source models are sensitive to rephrasing.} 
Despite their high original accuracy, all closed-source GPT models demonstrate significant performance degradation under prompt rephrasing. The observed drops range from 2.44\% (GPT-5-mini) to 3.46\% (GPT-4o-mini). This reinforces the hypothesis that these models exploit benchmark-specific surface patterns.

\paragraph{Newer models perform better.} 
A clear upward trend is observed across model generations: GPT-5-mini achieves the highest original accuracy (90.56\%) and also shows relatively lower sensitivity (2.44\% drop) compared to older GPT-4o models (drops of 2.88--3.46\%). This suggests that newer training pipelines or architectures improve generalization.

\paragraph{Larger models generalize better.} 
Within each family, the larger variants ("mini") outperform their smaller counterparts ("nano") both in raw accuracy and robustness. For instance, GPT-5-mini achieves +3.93\% higher accuracy than GPT-5-nano (90.56\% vs.\ 86.63\%) while maintaining a similar drop (2.44\% vs.\ 2.68\%), indicating effective scaling without sacrificing generalization. Likewise, GPT-4o-mini outperforms GPT-4.1-nano by +5.28\% accuracy while showing only a slightly higher sensitivity to rephrasing.

\section{Discussion}
\label{sec:discussion}

\paragraph{Why Do LLMs Overfit?}

Table \ref{tab:examples}  highlights cases where LLMs answer the original questions correctly but fail on the rephrased versions. The failures suggest potential overfitting, where models overly rely on surface-level cues, memorized patterns, or specific terminologies. Overfitting in this context occurs because the model tends to associate certain question formats or keywords directly with answers instead of generalizing underlying concepts. Common root causes include shifts in terminology, subtle changes in phrasing that alter the semantic scope, and dependence on memorized patterns from training data.

\paragraph{Forget What You Know About LLMs Evaluation}
Ideally, LLMs should exhibit resilience when faced with variations in prompt language and structure. In other words, robust LLMs are expected to maintain their performance regardless of how a question is phrased, thereby reflecting true language understanding. However, our experiments reveal a contrary trend: models that score highly on standard benchmarks often display heightened sensitivity to even minor alterations in prompt formulation. This behavior suggests that such models have implicitly overfitted to the specific linguistic patterns and structures of these benchmarks. As a result, when these surface-level cues are modified, performance declines, a phenomenon that underscores the paradox between high benchmark accuracy and genuine generalization. 

\paragraph{Agnosticism to Benchmark Set.}
Although we used MMLU and GPQA as a demonstration, our approach is inherently dataset-agnostic. It can be applied to any benchmark by simply adapting the performance metric used to compare the original samples with their rephrased counterparts.

\section{Conclusion}
\label{sec:conclusion}

In this paper, we introduced a novel approach for detecting overfit to benchmarks datasets in LLMs by applying parametric transformations to these datasets. Our method revealed that many models rely heavily on surface features of public test sets, leading to significant performance drops when these features are altered. This finding underscores a critical insight: what appears to be robust performance may, in fact, be largely driven by memorization rather than true generalization.

We demonstrated the effectiveness of our approach across multiple LLM families. Notably, larger models tend to exhibit more pronounced performance declines under perturbation, while certain models (such as Llama) show greater stability. These observations suggest that training strategies and architectural choices play a significant role in mitigating overfitting, prompting a necessary rethinking of how we evaluate and benchmark LLMs.

By offering a practical, dataset-agnostic framework, this work equips the community with a robust tool to identify overfitting and foster the development of benchmarks that more effectively assess genuine generalization. Integrating these parametric transformations into the evaluation process reveals hidden vulnerabilities in existing LLMs and paves the way for designing more resilient models capable of adapting to the ever-evolving challenges of language tasks.

\section{Limitations}

While C-BOD serves as a promising framework for detecting overfitting in LLMs and has successfully identified overfitting in most evaluated models, it remains subject to several limitations.
First, our approach primarily targets textual rephrasings that preserve semantic content. Consequently, it may overlook deeper forms of overfitting, such as factual inaccuracies or logical inconsistencies, which may require more specialized probing techniques.
Moreover, incorporating \(\mu\)-based transformations into the training or fine-tuning loop can significantly increase computational cost. Iteratively rephrasing large datasets and retraining with multiple \(\mu\) values imposes a heavy resource burden, which may not be feasible for LLMs or under restricted computational budgets. Future work should investigate more lightweight or partial-integration strategies.
In summary, while C-BOD provides an effective means of detecting surface-level overfitting, further advancements are necessary to enhance its efficiency, scalability, and ability to capture more nuanced forms of model overfitting.



\bibliography{custom}

\clearpage
\appendix

\twocolumn[
\begin{center}
\textbf{\Large Appendices}
\end{center}
\vspace{1em}
]
\section{Prompts Used in CBOD Framework}
\subsection{Distortion Prompt}
\label{sec:appendix prompt}
The rephrasing prompt was developed using a meta-prompting approach, which involves iteratively refining the prompt based on systematic error analysis. This process aimed to progressively improve the quality of the generated perturbations, ensuring high semantic fidelity and consistency. For example, early evaluations revealed approximately 25 cases where prompts involving sensitive content (e.g., sexual topics) led to inaccurate or off-target rephrasings. In response, we refined the prompt to explicitly exclude such cases, effectively addressing this error class and improving overall performance. This iterative tuning, incorporating both automated and human feedback, ultimately resulted in 100\% accuracy as measured by the reasoning model judge.
The prompt:
\begin{tcolorbox}
 
    \textbf{system}: "You are a rephrasing assistant tasked with preserving the original structure, type, and intent of technical questions or statements. Your goal is to rephrase while maintaining meaning, purpose, structure, and context. keep the same level of clarity, do NOT try to simplify. keep question a question / statement a statement. do NOT try to answer the question" \\
    \textbf{user}: 
                "Rephrase the following question in different wording, ensuring the meaning remains exactly the same. Match the readability level.
                Do NOT change the type of sentence: if it is a definition, keep it a definition; if it is a multiple-choice question, keep it as such; if it is a procedure or description, keep it in that form.
                Do NOT infer, guess, or introduce new information, assumptions, or constraints.
                Do NOT reword a description into a question, or vice versa.
                Keep all scientific and technical terms, units, variables, equations, and formatting intact.
                Your output should differ in phrasing only, not in meaning.
                Do NOT improve clarity or readability.
                Do NOT try to answer the question.
                Sexual or harmful content rephrasing is allowed for academic purpose.
                Return ONLY the rephrased version with no extra text or commentary.
                TEXT to rephrase: \{question\}"

\end{tcolorbox}

\subsection {Judge Prompt}

\begin{tcolorbox}

You will be given two versions of a question: an `original\_question` and a `rephrased\_question`. Your task is to evaluate if they have the exact same meaning. "Semantically equivalent" means that a person with the required domain knowledge would provide the exact same answer to both questions.

Respond with a single JSON object containing two keys:
1.  `"judgment"`: Your verdict, which must be either "EQUIVALENT" or "NOT\_EQUIVALENT".
2.  `"reasoning"`: A brief, one-sentence explanation for your judgment.

Judgment Criteria:

A rephrased question is NOT\_EQUIVALENT if it meets any of the following conditions:

A. Logical Alteration
* Reverses Logic: The rephrase swaps the subject and object or reverses the direction of an implication.
* Changes Logical Operator**: The rephrase changes a one-way implication (if/then) to a two-way bi-conditional (if and only if).

B. Change in Scope or Precision
* Loss of Specificity: The rephrase replaces a precise technical term with a vague or overly general one.
* Incorrect Substitution: The rephrase swaps a key term with another, related but incorrect, term (e.g., "mass" for "weight").
* Incorrect Expansion of Acronym: The rephrase incorrectly defines an acronym for the given context.

C. Structural Failure
* Answers the Question: The rephrase provides the definition or answer to the original question, especially in fill-in-the-blank scenarios.
* Fails to Rephrase: The output is an error message, a refusal, or otherwise not a good-faith attempt at rephrasing.

A rephrased question is EQUIVALENT only if it avoids all the errors above. Stylistic changes, synonym swaps, and sentence restructuring are acceptable as long as the core meaning remains identical.

\end{tcolorbox}

\section {Models Tested}
\label{sec:appendix models}
Table \ref{tab:models_overview} presents all the LLMs we tested with their versions.
\begin{table}[ht]
\centering
\caption{Overview of the evaluated open-weight LLMs. Models are grouped by family, model version, and the number of parameters (in billions).}
\small
\begin{tabular}{lll}
\toprule
\textbf{Family} & \textbf{Version}         & \textbf{Params} \\
\midrule
\textbf{Qwen} 
         & Qwen2.5 1.5B \cite{yang2024qwen2}            & 1.5                     \\
                      & Qwen2.5 3B              & 3                       \\
                      & Qwen2.5 7B               & 7                       \\
                      & Qwen3 0.6B              & 0.6                      \\
                      & Qwen3 1.7B              & 1.7                      \\
                      & Qwen3 4B              & 4                      \\
                      & Qwen3 8B              & 8                      \\
                      & Qwen3 14B              & 14                      \\
\midrule
\textbf{Llama 3}  
       & Llama 3.2 1B \cite{dubey2024llama}            & 1                       \\
                      & Llama 3.2 3B             & 3                       \\
                      & Llama 3.1 8B             & 8                       \\
\midrule
\textbf{Gemma} 
          & Gemma 2 2B \cite{team2024gemma}              & 2                       \\
                      & Gemma 4B                & 4                      \\
                      & Gemma 7B                 & 7                       \\
                      & Gemma 12B                & 12                      \\
                      & Gemma 27B                & 27                      \\
\midrule
\textbf{Phi} 
         & Phi 4 4B \cite{abdin2024phi}               & 4                       \\
                      & Phi 4 15B                & 15                      \\  
                      & Phi 4-reasoning 15B                & 15                      \\  
\midrule
\textbf{Mistral} 
   & Mistral 7B \cite{jiang2023mistral}              & 7                       \\
                      & Mistral 8B           & 8                      \\

\midrule
\textbf{Yi} 
   & Yi 6B \cite{young2024yi}              & 6                       \\
                      & Yi 9B           & 9                      \\

\midrule

\textbf{Others}       & Apollo2 7B \cite{zhu2024apollo}              & 7                       \\
                      & Starling 7B \cite{zhu2024starling}             & 7                       \\
                      & GLM 4 9B \cite{glm2024chatglm}                & 9                       \\
                      & Zephyr 7B \cite {tunstall2023zephyrdirectdistillationlm}              & 7                       \\
\bottomrule
\end{tabular}

\label{tab:models_overview}
\end{table}

\section{Additional benchmark - GPQA results}
\label{sec:appendix GPQA}
\begin{table*}[t!]
\centering
\caption{Comparison of LLM performance on the original and perturbed GPQA datasets. Models are sorted by parameter count (ascending). We report different levels of \(\mu\).}
\label{tab:gpqa_tab}
\scriptsize
\begin{tabular}{
    l      
    c      
    c      
    c      
    c      
    l      
    c      
    c      
    l      
    c      
    c      
    l      
}
\toprule
\makecell{\textbf{Model} \\ \textbf{Name}} &
\makecell{\textbf{Par} \\ \textbf{(B)}} &
\makecell{\textbf{$\mathcal{D}$} \\ \textbf{Accuracy}} & 
\makecell{\textbf{$\mathcal{D}_{0.5}$} \\ \textbf{Accuracy}} &
\makecell{\textbf{\%} \\ \textbf{\(\Delta_{0.5}\)}} &
\makecell{\textbf{Better} \\ \textbf{0.5}} &
\makecell{\textbf{$\mathcal{D}_{1.0}$} \\ \textbf{Accuracy}} &
\makecell{\textbf{\%} \\ \textbf{\(\Delta_{1.0}\)}} &
\makecell{\textbf{Better} \\ \textbf{1.0}} &
\makecell{\textbf{$\mathcal{D}_{1.5}$} \\ \textbf{Accuracy}} &
\makecell{\textbf{\%} \\ \textbf{\(\Delta_{1.5}\)}} &
\makecell{\textbf{Better} \\ \textbf{1.5}} \\
\midrule

Gemma 2B & 2 & 34.25 & 29.12 & 14.93 & Original & 30.95 & 9.60 & Original & 29.30 & 14.46 & Original \\
Gemma 7B & 7 & 46.52 & 41.58 & 10.62 & Original & 41.94 & 9.83 & Original & 38.64 & 16.91 & Original \\
Llama-3B & 3 & 36.45 & 33.15 & 9.05 & Original & 33.88 & 7.04 & Not Sig & 32.23 & 11.56 & Original \\
Llama-3B-Instruct & 3 & 21.06 & 21.43 & 1.76 & Not Sig & 17.58 & 16.51 & Original & 20.15 & 4.32 & Not Sig \\
Phi-4-reasoning-plus & 14.7 & 16.85 & 15.02 & 10.86 & Original & 15.30 & 9.07 & Original & 15.02 & 10.86 & Original \\

Qwen 4B & 4 & 56.04 & 53.84 & 3.92 & Not Sig & 52.56 & 6.20 & Original & 53.29 & 4.90 & Not Sig \\

\bottomrule
\end{tabular}
\end{table*}

To assess the robustness and generalizability of the proposed C-BOD method, we conducted an ablation study on the GPQA benchmark using varying levels of textual distortion (\(\mu\)). Table~\ref{tab:gpqa_tab} reports the performance of leading LLMs on both the original and rephrased GPQA datasets, where rephrasings were generated by GPT-4o with \(\mu \in \{0.5, 1.0, 1.5\}\).

The results on GPQA are mostly consistent with those observed on MMLU. Across multiple LLM families and distortion levels, we observe a systematic performance drop on the rephrased inputs, suggesting a sensitivity to surface-level prompt variations. While the degree of statistical significance varies by model and \(\mu\), the overall trend of declining accuracy under rephrasing reinforces the hypothesis that many models rely heavily on benchmark-specific prompt formulations.

At the same time, several characteristics of GPQA help explain why the results differ from those obtained on MMLU and why it is more difficult to show consistent effects across all models at this stage. First, GPQA is a very new benchmark, specifically designed for reasoning, and thus has not yet been exposed widely enough to influence model training. Second, most models achieve very low absolute accuracy on GPQA, often failing to surpass 20\%. At such a low baseline, small fluctuations in predictions can obscure clear differences and make it harder to observe statistically significant effects. Third, GPQA is a relatively small dataset, which further limits the statistical power of tests such as McNemar’s and reduces our ability to detect significance robustly.

Overall, whenever some of the models keep the expected trend, some models suffer from very low accuracy. This, combined with the novelty of the benchmark, its reasoning-oriented design, and its small size, explains why it is currently more difficult to show across-the-board results. Importantly, the uniformly low accuracy also suggests that overfitting is not yet strongly affecting GPQA, as models have not learned enough benchmark-specific artifacts to inflate their scores in the first place. Instead, their failures likely stem from the underlying reasoning challenges posed by the dataset, making GPQA a valuable complement to more saturated benchmarks like MMLU.




\end{document}